\documentclass[10pt,twocolumn,letterpaper]{article}

\usepackage{iccv}
\usepackage{times}
\usepackage{epsfig}
\usepackage{graphicx}
\usepackage{amsmath}
\usepackage{amssymb}
\usepackage{amsfonts}
\usepackage{booktabs}
\usepackage{color}

\usepackage[pagebackref=true,breaklinks=true,letterpaper=true,colorlinks,bookmarks=false]{hyperref}

 \iccvfinalcopy 

\def\iccvPaperID{355} 

\ificcvfinal\fi
\begin{document}

\title{Unsupervised Learning of Visual Representations using Videos}

\author{Xiaolong Wang, Abhinav Gupta\\
Robotics Institute, Carnegie Mellon University\\
}

\maketitle

\begin{abstract}
Is strong supervision necessary for learning a good visual representation? Do we really need millions of semantically-labeled images to train a Convolutional Neural Network (CNN)? In this paper, we present a simple yet surprisingly powerful approach for unsupervised learning of CNN. Specifically, we use hundreds of thousands of unlabeled videos from the web to learn visual representations. Our key idea is that visual tracking provides the supervision. That is, two patches connected by a track should have similar visual representation in deep feature space since they probably belong to the same object or object part. We design a Siamese-triplet network with a ranking loss function to train this CNN representation. Without using a single image from ImageNet, just using 100K unlabeled videos and the VOC 2012 dataset, we train an ensemble of unsupervised networks that achieves $~52\%$ mAP (no bounding box regression). This performance comes tantalizingly close to its ImageNet-supervised counterpart, an ensemble which achieves a mAP of $~54.4\%$. We also show that our unsupervised network can perform competitively in other tasks such as surface-normal estimation.
\end{abstract}

\vspace{-0.2in}
\section{Introduction}
\vspace{-0.07in}
What is a good visual representation and how can we learn it? At the start of this decade, most computer vision research focused on ``what'' and used hand-defined features such as SIFT~\cite{Lowe2004} and HOG~\cite{Dalal05} as the underlying visual representation. Learning was often the last step where these low-level feature representations were mapped to semantic/3D/functional categories. However, the last three years have seen the resurgence of learning visual representations directly from pixels themselves using the deep learning and Convolutional Neural Networks (CNNs)~\cite{lecun90,alexnet12,jia2014caffe}. At the heart of CNNs is a completely supervised learning paradigm. Often millions of examples are first labeled using Mechanical Turk followed by data augmentation to create tens of millions of training instances. CNNs are then trained using gradient descent and back propagation. But one question still remains: is strong-supervision necessary for training these CNNs? Do we really need millions of semantically-labeled images to learn a good representation? It seems humans can learn visual representations using little or no semantic supervision but our approaches still remain completely supervised.

\begin{figure}
\includegraphics[width=0.5\textwidth]{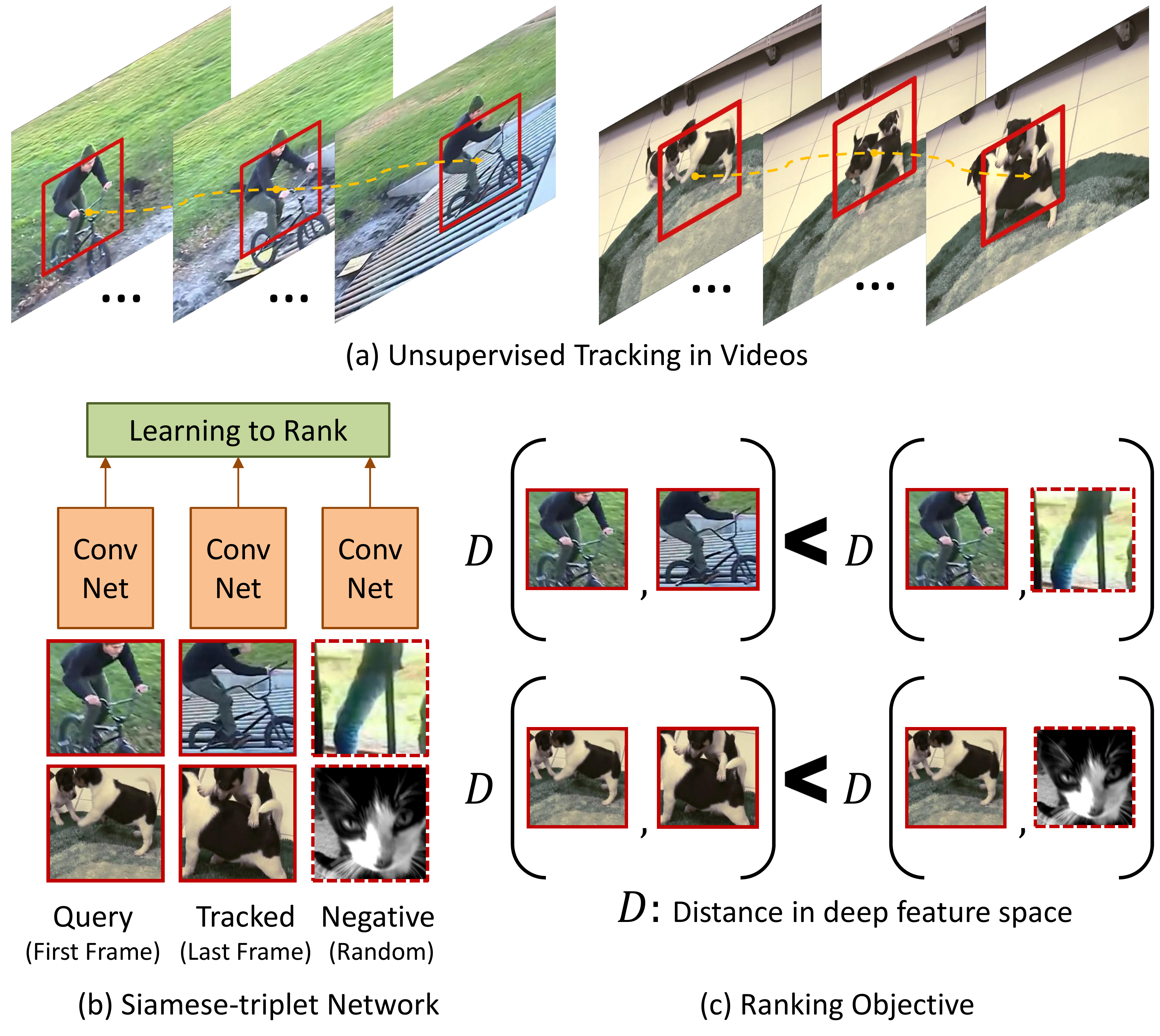}
\caption{\label{fig:teaser}
Overview of our approach. (a) Given unlabeled videos, we perform unsupervised tracking on the patches in them. (b) Triplets of patches including query patch in the initial frame of tracking, tracked patch in the last frame, and random patch from other videos are fed into our siamese-triplet network for training. (c) The  learning objective: Distance between the query and tracked patch in feature space should be smaller than the distance between query and random patches.}
\vspace{-0.2in}
\end{figure}

In this paper, we explore the alternative: how we can exploit the unlabeled visual data on the web to train CNNs (e.g. AlexNet~\cite{alexnet12})? In the past, there have been several attempts at unsupervised learning using millions of static images~\cite{quocle12,ruslanDBM12} or frames extracted from videos~\cite{ngVideo12,taylor10,Mobahi09}. The most common architecture used is an auto-encoder which learns representations based on its ability to reconstruct the input images~\cite{Olshausen1997,Bengio07,vincent08,Ranzato07}. While these approaches have been able to automatically learn V1-like filters given unlabeled data, they are still far away from supervised approaches on tasks such as object detection. So, what is the missing link? We argue that static images themselves might not have enough information to learn a good visual representation. But what about videos? Do they have enough information to learn visual representations? In fact, humans also learn their visual representations not from millions of static images but years of dynamic sensory inputs. Can we have similar learning capabilities for CNNs?

We present a simple yet surprisingly powerful approach for unsupervised learning of CNNs using hundreds of  thousands of unlabeled videos from the web. Visual tracking is one of the first capabilities that develops in infants and often before semantic representations are learned\footnote{http://www.aoa.org/patients-and-public/good-vision-throughout-life/childrens-vision/infant-vision-birth-to-24-months-of-age}. Taking a leaf from this observation, we propose to exploit visual tracking for learning CNNs in an unsupervised manner. Specifically, we track millions of ``moving'' patches in hundreds of thousands of videos. Our key idea is that two patches connected by a track should have similar visual representation in  deep feature space since they probably belong to the same object. We design a Siamese-triplet network with ranking loss function to train the CNN representation. This ranking loss function enforces that in the final deep feature space the first frame patch should be much closer to the tracked patch than any other randomly sampled patch. We demonstrate the strength of our learning algorithm using extensive experimental evaluation. Without using a single image from ImageNet, just 100K unlabeled videos and VOC 2012 dataset, we train an ensemble of AlexNet networks that achieves $52\%$ mAP (no bounding box regression). This performance is similar to its ImageNet-supervised counterpart, an ensemble which achieves $54.4\%$ mAP. We also show that our network trained using unlabeled videos achieves similar performance to its completely supervised counterpart on other tasks such as surface normal estimation. We believe this is the first time an unsupervised-pretrained CNN has been shown so competitive; that too on varied datasets and tasks. Specifically for VOC, we would like to put our results in context: this is the best results till-date by using only PASCAL-provided annotations (next best is scratch at $44\%$).

\vspace{-0.1mm}
\section{Related Work}
\vspace{-0.5mm}
Unsupervised learning of visual representations has a rich history starting from original auto-encoders work of Olhausen and Field~\cite{Olshausen1997}. Most of the work in this area can be broadly divided into three categories. The first class of algorithms focus on learning generative models with strong priors~\cite{hinton1995,sudderth2006}. These algorithms essentially capture co-occurrence statistics of features. The second class of algorithms use manually defined features such as SIFT or HOG and perform clustering over training data to discover semantic classes~\cite{sivic2005,russell2006}. Some of these recent algorithms also focus on learning mid-level representations rather than discovering semantic classes themselves~\cite{Singh2012, Doersch2013, Doersch2014}.
The third class of algorithms and more related to our paper is unsupervised learning of visual representations from the pixels themselves using deep learning approaches~\cite{hinton06,quocle12,ruslanDBM12,nyuPedstrain13,Honglak09,RoBM2012, shapeRBM2012,pluo12,Bengio13,vincent08,CarlUnsup2015}. Starting from the seminal work of Olhausen and Field~\cite{Olshausen1997}, the goal is to learn visual representations which are (a) sparse and (b) reconstructive. Olhausen and Field~\cite{Olshausen1997} showed that using this criteria they can learn V1-like filters directly from the data. However, this work only focused on learning a single layer. This idea was extended by Hinton and Salakhutdinov~\cite{hinton06} to train a deep belief network in an unsupervised manner via stacking layer-by-layer RBMs. Similar to this, Bengio et al.~\cite{Bengio07} investigated stacking of both RBMs and autoencoders. As a next step, Le et al.~\cite{quocle12} scaled up the learning of multi-layer autoencoder on large-scale unlabeled data. They demonstrated that although the network is trained in an unsupervised manner, the neurons in high layers can still have high responses on semantic objects such as human heads and cat faces. Sermanet et al.~\cite{nyuPedstrain13} applied convolutional sparse coding to pre-train the model layer-by-layer in unsupervised manner. The model is then fine-tuned for  pedestrian detection. In a contemporary work, Doersch et al.~\cite{CarlUnsup2015} explored to use spatial context as a cue to perform unsupervised learning for CNNs.

However, it is not clear if static images is the right way to learn visual representations. Therefore, researchers have started focusing on learning feature representations using videos~\cite{Foldiak1991, Wiskott2002, quocleCVPR11, UnsupLSTM15, ngVideo12,nyuUnsup15,taylor10,Mobahi09,DavidVideo2010}. Early work such as~\cite{ngVideo12} focused on inclusion of constraints via video to autoencoder framework. The most common constraint is enforcing learned representations to be temporally smooth. Similar to this, Goroshin et al.~\cite{nyuUnsup15} proposed to learn auto-encoders based on the slowness prior. Other approaches such as Taylor et al.~\cite{taylor10} trained convolutional gated RBMs to learn latent representations from pairs of successive images. This was extended in a recent work by Srivastava et al.~\cite{UnsupLSTM15} where they proposed to learn a LSTM model in an unsupervised manner to predict future frames.

Finally, our work is also related to metric learning via deep networks~\cite{JWRank14, MCNN15,Siamese05, Hadsell06, Gong14, Hoffer15,Paul2015}. For example, Chopra et al.~\cite{Siamese05} proposed to learn convolutional networks in a siamese architecture for face verification. Wang et al.~\cite{JWRank14} introduced a deep triplet ranking network to learn fine-grained image similarity.  Zhang et al.~\cite{Zhang-TIP15} optimized the max-margin loss on triplet units to learn deep hashing function for image retrieval. However, all these methods required labeled data. Our work is also related to~\cite{babylearning14}, which used CNN pre-trained on ImageNet classification and detection dataset as initialization, and performed semi-supervised learning in videos to tackle object detection in target domain. However, in our work, we propose an unsupervised approach instead of semi-supervised algorithm.

\vspace{-2.6mm}
\section{Overview}
\vspace{-2.4mm}
Our goal is to train convolutional neural networks using hundreds of thousands of unlabeled videos from the Internet. We follow the AlexNet architecture to design our base network. However, since we do not have labels, it is not clear what should be the loss function and how  we should optimize it. But in case of videos, we have another supervisory information: time. For example, we all know that the scene does not change drastically within a short time in a video and same object instances appear in multiple frames of the video. So, how do we exploit this information to train a CNN-based representation?

We sample millions of patches in these videos and track them over time. Since we are tracking these patches, we know that the first and last tracked frames  correspond to the same instance of the moving object or object part. Therefore, any visual representation that we learn should keep these two data points close in the feature space. But just using this constraint is not sufficient: all points can be mapped to a single point in feature space. Therefore, for training our CNN, we sample a third patch which creates a triplet. For training, we use a loss function~\cite{JWRank14} that enforces that the first two patches connected by tracking are closer in feature space than the first one and a random one.

Training a network with such triplets converges fast since the task is easy to overfit to. One way is to increase the number of training triplets. However, after initial convergence most triplets satisfy the loss function and therefore back-propagating gradients using such triplets is inefficient. Instead, analogous to hard-negative mining, we select the third patch from multiple patches that violates the constraint (loss is maximum). Selecting this patch leads to more meaningful gradients for faster learning.

\vspace{-1.7mm}
\section{Patch Mining in Videos}
\vspace{-1.7mm}

\begin{figure}
\includegraphics[width=0.48\textwidth]{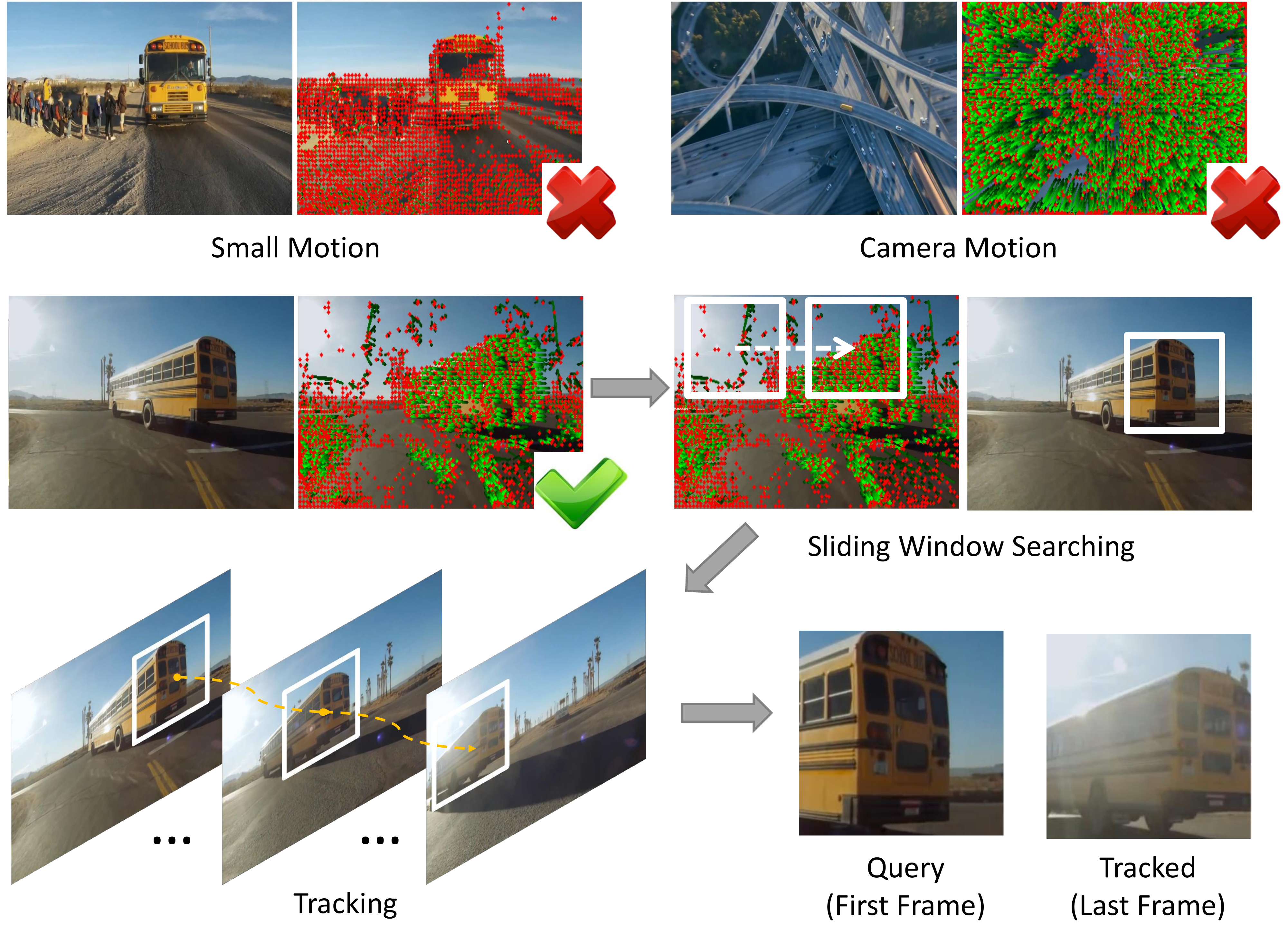}
\centering
\caption{\label{fig:track}
Given the video about buses (the ``bus'' label are not utilized), we perform IDT on it. \textcolor{red}{red} points represents the SURF feature points, \textcolor{green}{green} represents the trajectories for the points. We reject the frames with small and large camera motions (top pairs). Given the selected frame, we  find the bounding box containing most of the moving SURF points. We then perform tracking. The first and last frame of the track provide pair of  patches for training CNN. }
\vspace{-0.1in}
\end{figure}

\begin{figure*}
\includegraphics[width=6.8in]{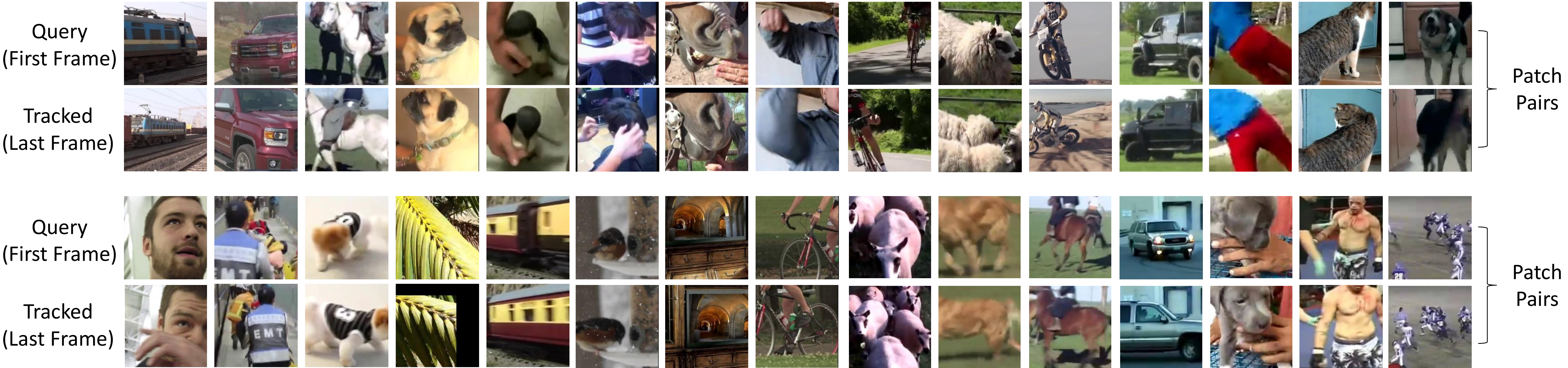}
\caption{\label{fig:mine}
Examples of patch pairs we obtain via patch mining in the videos. }
\vspace{-0.13in}
\end{figure*}

Given a video, we want to extract patches of interest (patches with motion in our case) and track these patches to create training instances. One obvious way to find patches of interest is to compute optical flow and use the high magnitude flow regions. However, since YouTube videos are noisy with a lot of camera motion, it is hard to localize moving objects using simple optical flow magnitude vectors. Thus we follow a two-step approach: in the first step, we obtain SURF~\cite{Bay06} interest points and use Improved Dense Trajectories (IDT)~\cite{IDT13} to obtain motion of each SURF point. Note that since  IDT applies a homography estimation (video stabilization) method, it reduces the problem caused by camera motion. Given the trajectories of SURF interest points, we classify these points as moving if the flow magnitude is more than 0.5 pixels. We also reject frames if (a) very few ($<25\%$) SURF interest points are classified as moving because it might be just noise; (b) majority of SURF interest points ($>75\%$) are classified as moving as it corresponds to moving camera. Once we have extracted moving SURF interest points, in the second step, we find the best bounding box such that it contains most of the moving SURF points. The size of the bounding box is set as $h \times w$, and we perform sliding window with it in the frame. We take the bounding box which contains the most number of moving SURF interest points as the interest bounding box. In the experiment, we set $h=227, w=227$ in the frame with size $448 \times 600$. Note that these patches might contain objects or part of an object as shown in Figure~\ref{fig:track}.

\noindent \textbf{Tracking.} Given the initial bounding box, we perform tracking using the KCF tracker~\cite{Tracker15}. After tracking along 30 frames in the video, we obtain the second patch. This patch acts as the similar patch to the query patch in the triplet. Note that the KCF tracker does not use any supervised information except for the initial bounding box.

\vspace{-2.0mm}
\section{Learning Via Videos}
\vspace{-1.6mm}
In the previous section, we discussed how we can use tracking to generate pairs of patches. We use this procedure to generate millions of such pairs (See Figure~\ref{fig:mine} for examples of pairs of patches mined). We now describe how we use these as training instances for our visual representation learning.

\vspace{-1.0mm}
\subsection{Siamese Triplet Network}
\begin{figure}[b]
\includegraphics[width=0.5\textwidth]{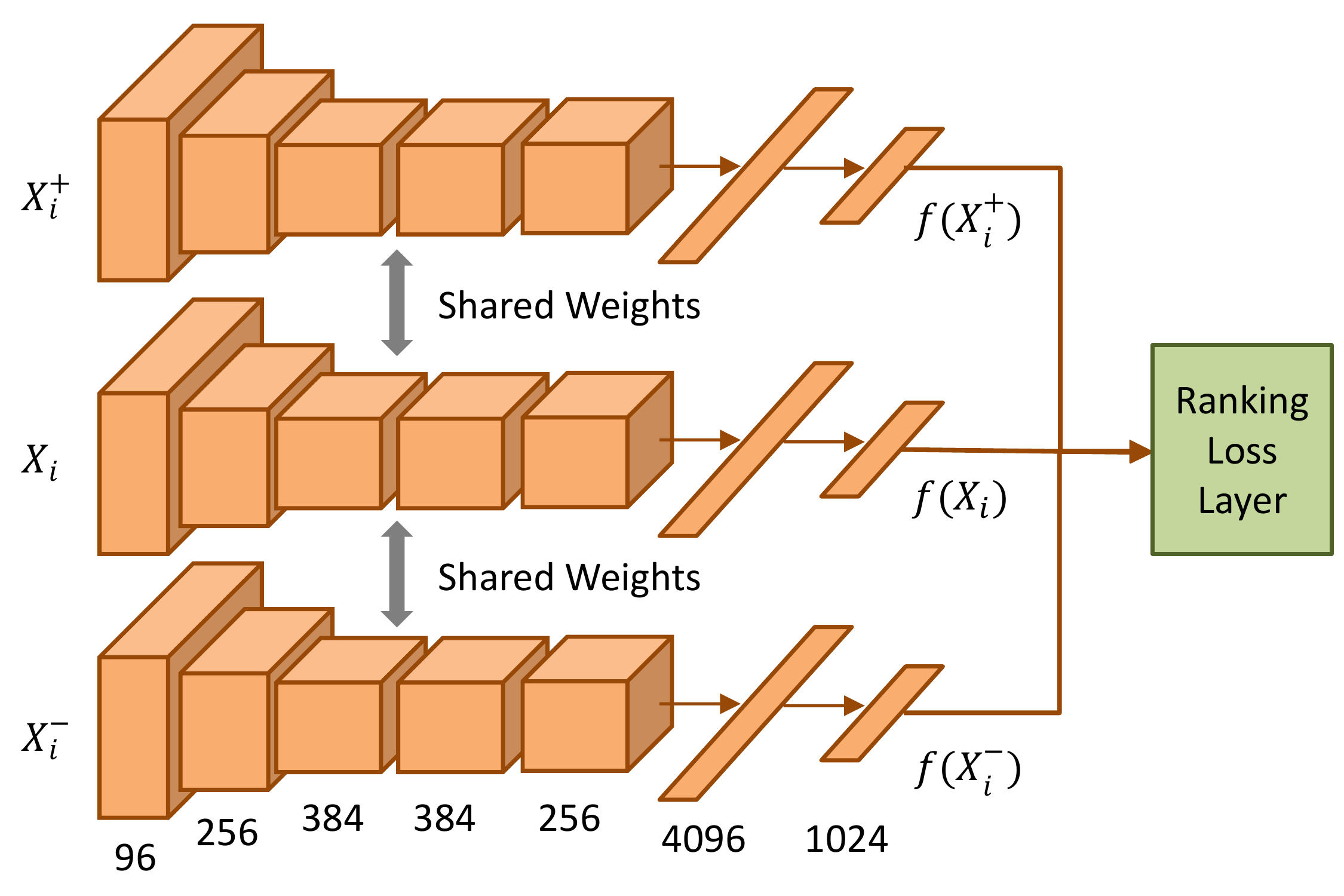}
\caption{\label{fig:triplet}
Siamese-triplet network. Each base network in the Siamese-triplet network share the same architecture and parameter weights. The architecture is rectified from AlexNet by using only two fully connected layers. Given a triplet of training samples, we obtain their features from the last layer by forward propagation and compute the ranking loss.}
\vspace{-0.15in}
\end{figure}

\vspace{-1.2mm}
Our goal is to learn a feature space such that the query patch is closer to the tracked patch as compared to any other randomly sampled patch. To learn this feature space we design a Siamese-triplet network. A Siamese-triplet network consist of three base networks which share the same parameters (see Figure~\ref{fig:triplet}). For our experiments, we take the image with size $227 \times 227 $ as input. The base network is based on the AlexNet architecture~\cite{alexnet12} for the convolutional layers. Then we stack two fully connected layers on the pool5 outputs, whose neuron numbers are 4096 and 1024 respectively. Thus the final output of each single network is $1024$ dimensional feature space $f(\cdot)$. We define the loss function on this feature space.

\vspace{-1.0mm}
\subsection{Ranking Loss Function}
\vspace{-1.0mm}
Given the set of patch pairs $\mathbb{S}$ sampled from the video, we propose to learn an image similarity model in the form of CNN. Specifically, given an image $X$ as an input for the network, we can obtain its feature in the final layer as $f(X)$. Then, we define the distance of two image patches $X_1, X_2$ based on the cosine distance in the feature space as,

\vspace{-2mm}
{\footnotesize
\begin{eqnarray}\label{eq:cos_dis}
D(X_1, X_2) = 1 - \frac{f(X_1) \cdot f(X_2)} {\|f(X_1)\| \|f(X_2)\| }.
\end{eqnarray}
}
\vspace{-2.5mm}

We want to train a CNN to obtain feature representation $f(\cdot)$, so that the distance between query image patch and the tracked patch is small and the distance between query patch and other random patches is encouraged to be larger. Formally, given the patch set $\mathbb{S}$, where $X_i$ is the original query patch (first patch in tracked frames),
$X_i^{+}$ is the tracked patch and $X_i^{-}$ is a random patch from a different video, we want to enforce $D(X_i, X_i^{-}) > D(X_i, X_i^{+})$. Therefore, the loss of our ranking model is defined by hinge loss as,

\vspace{-2.5mm}
{\footnotesize
\begin{eqnarray}\label{eq:cos_loss}
L(X_i, X_i^{+}, X_i^{-}) =  \max\{ 0,  D(X_i, X_i^{+}) - D(X_i, X_i^{-}) + M \},
\end{eqnarray}
}
where $M$ represents the gap parameters between two distances. We set $M = 0.5$ in the experiment. Then our objective function for training can be represented as,

\vspace{-3mm}
{\footnotesize
\begin{eqnarray}\label{eq:obj}
\min_{W} \frac{\lambda}{2} \parallel W  \parallel_{2}^{2} + \sum_{i=1}^{N}  \max\{ 0,  D(X_i, X_i^{+}) - D(X_i, X_i^{-}) + M \},
\end{eqnarray}
}
where $W$ is the parameter weights of the network, i.e., parameters for function $f(\cdot)$. $N$ is the number of the triplets of samples. $\lambda$ is a constant representing weight decay, which is set to $\lambda = 0.0005$.

\subsection{Hard Negative Mining for Triplet Sampling}
\vspace{-0.5mm}
One non-trivial part for learning to rank is the process of selecting negative samples. Given a pair of similar images $X_i, X_i^{+}$, how can we select the patch $X_i^{-}$, which is a negative match to $X_i$, from the large pool of patches? Here we first select the negative patches randomly, and then find hard examples (in a process analogous to hard negative mining).

\textbf{Random Selection:} During learning, we perform mini-batch Stochastic Gradient Descent (SGD). For each $X_i, X_i^{+}$, we randomly sample $K$ negative matches in the same batch $\mathbb{B}$, thus we have $K$ sets of triplet of samples.  For every triplet of samples, we calculate the gradients over three of them respectively and perform back propagation. Note that we shuffle all the images randomly after each epoch of training, thus the pair of patches $X_i, X_i^{+}$ can look at different negative matches each time.

\textbf{Hard Negative Mining:} While one can continue to sample random patches for  creating the triplets, it is more efficient to search the negative patches smartly. After 10 epochs of training using negative data selected randomly, we want to make the problem harder to get more robust feature representations. Analogous to hard-negative mining procedure in SVM, where gradient descent learning is only performed on hard-negatives (not all possible negative), we search for the negative patch such that the loss is maximum and use that patch to compute and back propagate gradients.

Specifically, the sampling of negative matches is similar as random selection before, except that this time we select according to the loss(Eq.~\ref{eq:cos_loss}). For each pair $X_i, X_i^{+}$, we calculate the loss of all other negative matches in batch $\mathbb{B}$, and select the top $K$ ones with highest losses. We apply the loss on these $K$ negative matches as our final loss and calculate the gradients over them. Since the feature of each sample is already computed after the forward propagation, we only need to calculate the loss over these features, thus the extra computation for hard negative mining is very small. For the experiments, we use $K=4$. Note that while some of the negatives might be semantically similar patches, our embedding constraint only requires same instance examples to be closer than category examples (which can be closer than other negatives in the space).

\begin{figure}
\includegraphics[width=0.44\textwidth]{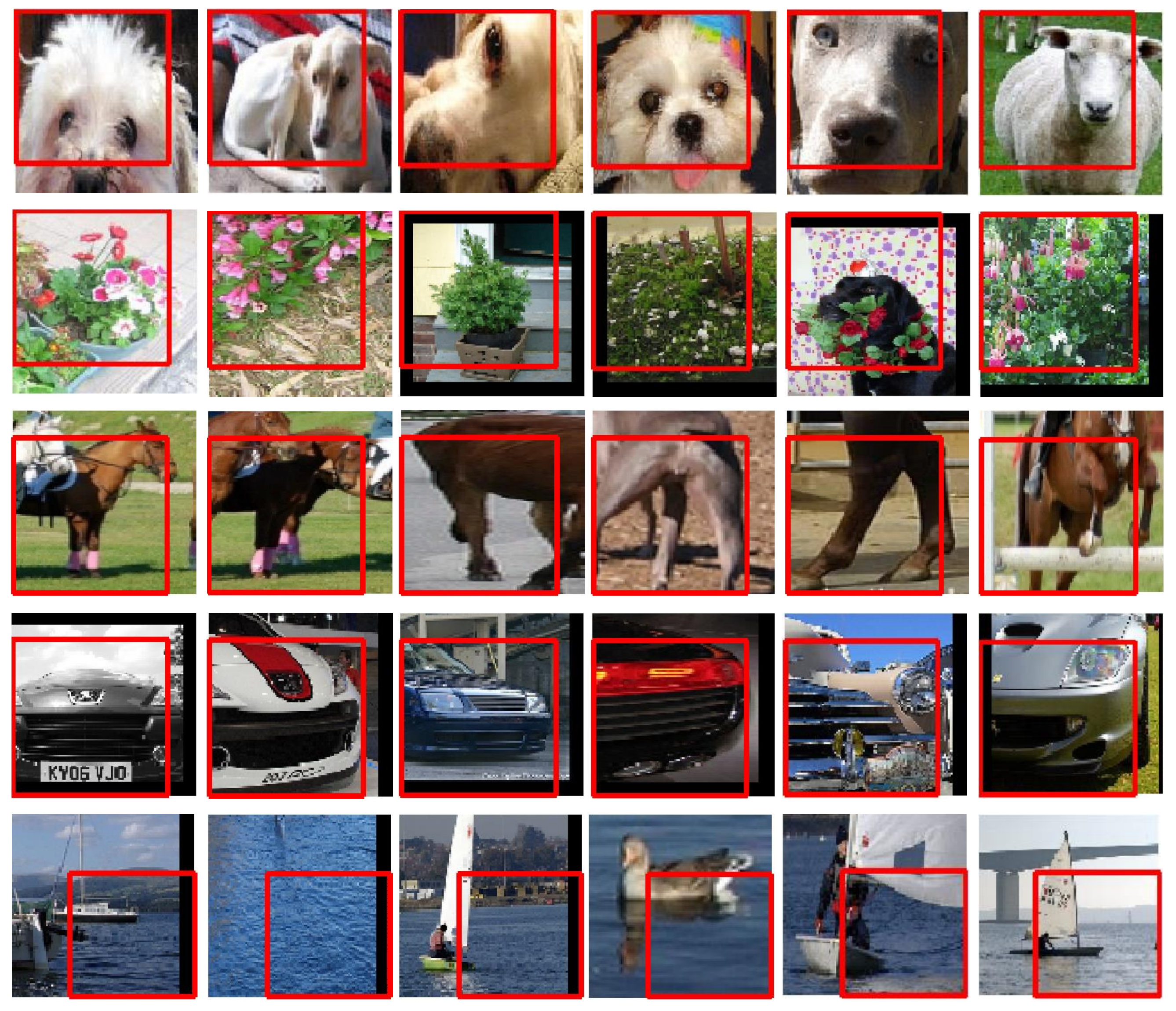}
\centering
\caption{\label{fig:activation}
Top response regions for the pool5 neurons of our unsupervised-CNN. Each row shows top response of one neuron.}
\vspace{-0.2in}
\end{figure}

\vspace{-1.0mm}
\subsection{Adapting for Supervised Tasks}
\vspace{-1.5mm}
Given the CNN learned by using unsupervised data, we want to transfer the learned representations to the tasks with supervised data. In our experiments, we apply our model to two different tasks including object detection and surface normal estimation. In both tasks we take the base network from our Siamese-triplet network and adjust the fully connected layers and outputs accordingly. We introduce two ways to fine-tune and transfer the information obtained from unsupervised data to supervised learning.

One straight forward approach is directly applying our ranking model as a pre-trained network for the target task. More specifically, we use the parameters of the convolutional layers in the base network of our triplet architecture as initialization for the target task. For the fully connected layers, we initialize them randomly. This method of transferring feature representation is very similar to the approach applied in RCNN~\cite{girshick14rcnn}. However, RCNN uses the network pre-trained with ImageNet Classification data. In our case, the unsupervised ranking task is quite different from object detection and surface normal estimation. Thus, we need to adapt the learning rate to the fine-tuning procedure introduced in RCNN. We start with the learning rate with $\epsilon =0.01$ instead of $0.001$ and set the same learning rate for all layers. This setting is crucial since we want the pre-trained features to be used as initialization of supervised learning, and adapting the features to the new task.

In this paper, we explore one more approach to transfer/fine-tune the network. Specifically, we note that there might be more juice left in the millions of unsupervised training data (which could not be captured in the initial learning stage). Therefore, we use an iterative fine-tuning scheme. Given the initial unsupervised network, we first fine-tune using the PASCAL VOC data. Given the new fine-tuned network, we use this network to re-adapt to ranking triplet task. Here we again transfer convolutional parameters for re-adapting. Finally, this re-adapted network is fine-tuned on the VOC data yielding a better trained model. We show in the experiment that this circular approach gives improvement in performance. We also notice that after two iterations of this approach the network converges.

\vspace{-1.0mm}
\subsection{Model Ensemble}
\vspace{-1.5mm}
We proposed an approach to learn CNNs using unlabeled videos. However, there is absolutely no limit to generating training instances and pairs of tracked patches (YouTube has more than billions of videos). This opens up the possibility of training multiple CNNs using different sets of data. Once we have trained these CNNs, we append the fc7 features from  each of these CNNs to train the final SVM. Note that the ImageNet trained models also provide initial boost for adding more networks (See Table~\ref{tab:VOCAP}).

\vspace{-1.0mm}
\subsection{Implementation Details}
\vspace{-1.5mm}
We apply mini-batch SGD in training. As the 3 networks share the same parameters, instead of inputting 3 samples to the triplet network, we perform the forward propagation for the whole batch by a single network and calculate the loss based on the output feature. Given a pair of patches $X_i, X_i^{+}$, we randomly select another patch $X_i^{-} \in \mathbb{B}$ which is extracted in a different video from $X_i, X_i^{+}$. Given their features from forward propagation $f(X_i), f(X_i^{+}), f(X_i^{-})$, we can compute the loss as Eq.~\ref{eq:cos_loss}.

For unsupervised learning, we download 100K videos from YouTube using the URLs provided by~\cite{babylearning14}.
\cite{babylearning14} used thousands of keywords to retrieve videos from YouTube. Note we drop the labels associated with each video. By performing our patch mining method on the videos, we obtain 8 million image patches. We train three different networks separately using 1.5M, 5M and 8M training samples. We report numbers based on these three networks. To train our siamese-triplet networks, we set the batch size as $|\mathbb{B}| = 100$, the learning rate starting with $\epsilon_0 =0.001$. We first train our network with random negative samples at  this learning rate for 150K iterations, and then we apply hard negative mining based on it. For training on 1.5M patches, we reduce the learning rate by a factor of 10 at every 80K iterations and train for 240K iterations. For training on 5M and 8M patches, we reduce the learning rate by a factor of 10 at every 120K iterations and train for 350K iterations.

\vspace{-2mm}
\section{Experiments}
\begin{figure}
\includegraphics[width=0.48\textwidth]{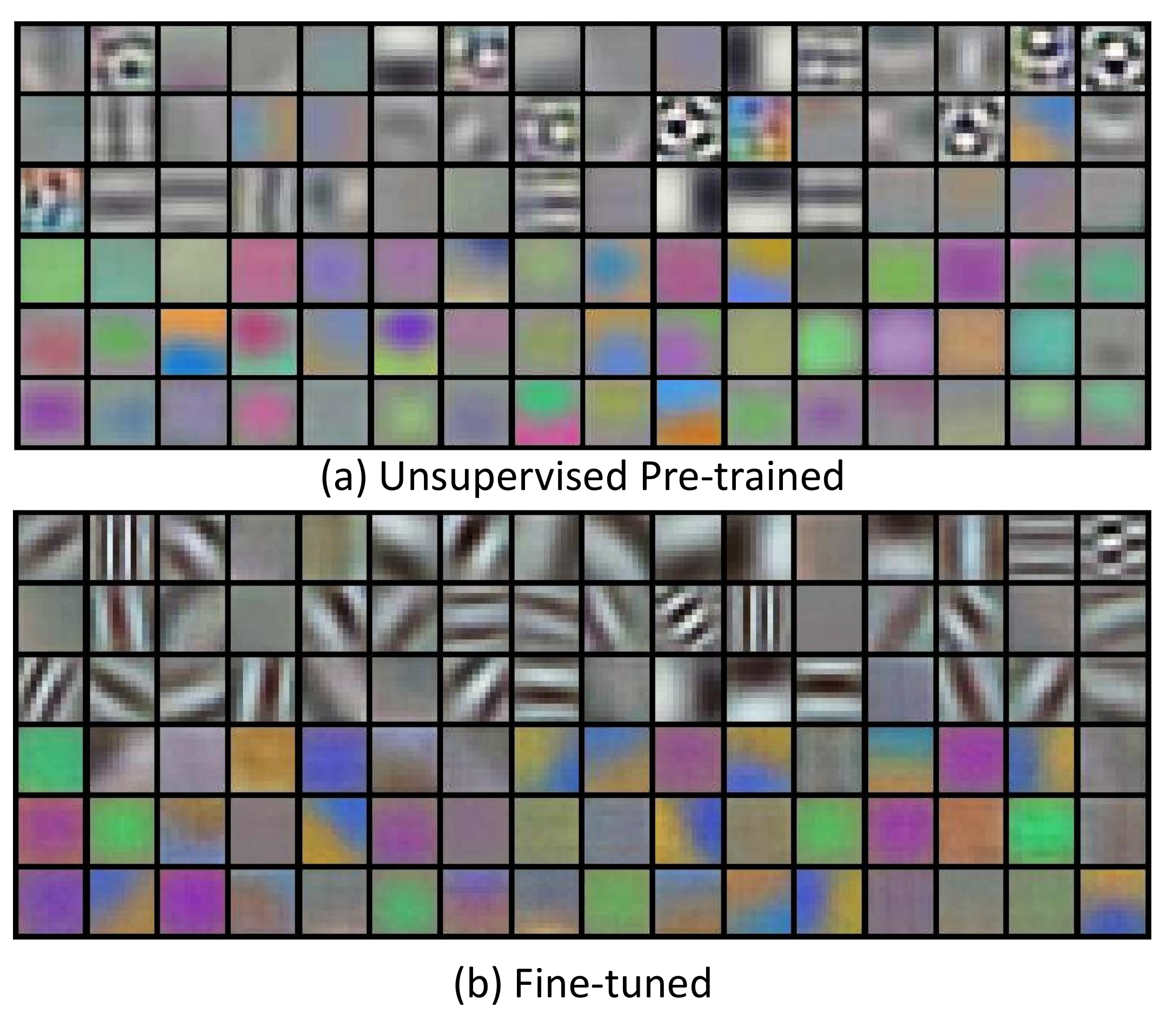}
\centering
\caption{\label{fig:filters}
Conv1 filters visualization. (a) The filters of the first convolutional layer of the siamese-triplet network trained in unsupervised manner. (b) By fine-tuning the unsupervised pre-trained network on PASCAL VOC 2012, we obtain sharper filters.}
\vspace{-0.15in}
\end{figure}

\vspace{-1.5mm}
We demonstrate the quality of our learned visual representations with qualitative and quantitative experiments. Qualitatively, we show the convolutional filters learned in layer 1 (See Figure~\ref{fig:filters}). Our learned filters are similar to V1 though not as strong. However, after fine-tuning on PASCAL VOC 2012, these filters become quite strong.  We also show that the underlying representation learns a reasonable nearness metric by showing what the units in Pool5 layers represent (See Figure~\ref{fig:activation}). Ignoring boundary effects, each pool5 unit has a receptive field of $195 \times 195$ pixels in the original $227 \times 227$ pixel input. A central pool5 unit has a nearly global view, while one near the edge has a smaller, clipped support. Each row displays top 6 activations for a pool5 unit. We have chosen 5 pool5 units for visualization. For example, the first neuron represents animal heads, second represents potted plant, etc. This visualization indicates the nearness metric learned by the network since each row corresponds to similar firing patterns inside the CNN. Our unsupervised networks are available for download.

\begin{figure*}
\includegraphics[width=0.95\textwidth]{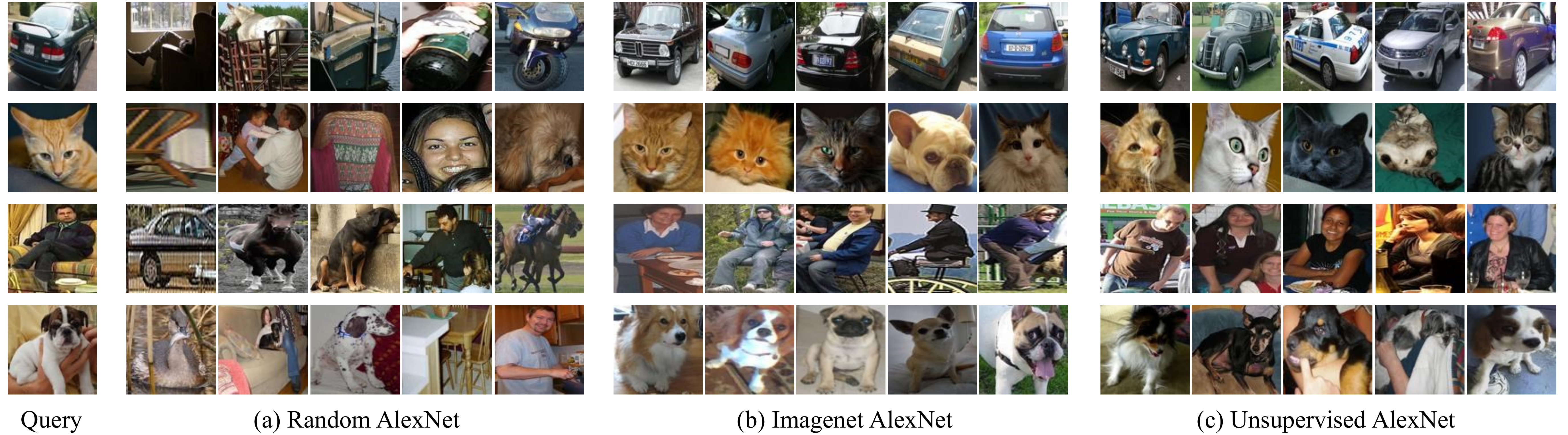}
\centering
\caption{\label{fig:nearest_neighbor}
Nearest neighbors results. Given the query object from VOC 2012 val, we retrieve the NN from VOC 2012 train via calculating the cosine distance on pool5 feature space. We compare the results of 3 different models: (a) AlexNet with random parameters; (b) AlexNet trained with Imagenet data; (c) AlexNet trained using our unsupervised method on 8M data.}
\vspace{-0.1in}
\end{figure*}

\vspace{-0.05in}
\subsection{Unsupervised CNNs without Fine-tuning}
\vspace{-0.08in}
First, we demonstrate that the unsupervised-CNN representation learned using videos (without fine-tuning) is reasonable. We perform Nearest Neighbors (NN) using ground-truth (GT) windows in VOC 2012 val set as query. The retrieval-database consists of all selective search windows  (more than 0.5 overlap with GT windows) in VOC 2012 train set. See Figure~\ref{fig:nearest_neighbor} for qualitative results. Our unsupervised-CNN is far superior to a random AlexNet architecture and the results are quite comparable to AlexNet trained on ImageNet.

Quantitatively, we measure the retrieval rate by counting number of correct retrievals in top-K (K=20) retrievals. A retrieval is correct if the semantic class for retrieved patch and query patch are the same. Using our unsupervised-CNN (Pool5 features) without fine-tuning and cosine distance, we obtain 40\% retrieval rate. Our performance is significantly better as compared  to 24\% by ELDA~\cite{ELDA2012} on HOG and 19\% by AlexNet with random parameters (our initialization). This clearly demonstrates our unsupervised network learns a good visual representation compared to a random parameter CNN. As a baseline, ImageNet CNN performs 62\% (but note it already learns on semantics).

We also evaluate our unsupervised-CNN without fine-tuning for scene classification task on MIT Indoor 67~\cite{MITIndoor2009}. We train a linear classifier using softmax loss. Using pool5 features from unsupervised-CNN without fine-tuning gives 41\% classification accuracy compared to 21\% for GIST+SVM and 16\% for random AlexNet. ImageNet-trained AlexNet has 54\% accuracy. We also provide object detection results without fine-tuning in the next section.

\vspace{-0.05in}
\subsection{Unsupervised CNNs with Fine-tuning}
\vspace{-0.06in}
Next, we evaluate our approach by transferring the feature representation learned in unsupervised manner to the tasks with labeled data. We focus on two challenging problems: object detection and surface normal estimation.

\vspace{-0.16in}
\subsubsection{Object Detection}
\vspace{-0.06in}
For object detection, we perform our experiments on PASCAL VOC 2012 dataset~\cite{PASCAL12}. We follow the detection pipeline introduced in RCNN~\cite{girshick14rcnn}, which borrowed the CNNs pre-trained on other datasets and fine-tuned on it with the VOC data. The fine-tuned CNN was then used to extract features followed by training SVMs for each object class. However, instead of using ImageNet pre-trained network as initialization in RCNN, we use our unsupervised-CNN. We fine-tune our network with the trainval set (11540 images) and train SVMs with them. Evaluation is performed in the standard test set (10991 images).

\begin{table*}
\centering
\caption{{\footnotesize mean Average Precision (mAP) on VOC 2012. ``external'' column shows the number of patches used to pre-train unsupervised-CNN.}}

\renewcommand{\arraystretch}{1.2}
\renewcommand{\tabcolsep}{1.2mm}
\resizebox{\linewidth}{!}{
\begin{tabular}{@{}l|l|r*{19}{c}|c@{}}
\hline
\textbf{VOC 2012 test}   &external & aero      & bike      & bird      & boat      & bottle     & bus        & car        & cat        & chair      & cow        & table      & dog        & horse      & mbike      & person     & plant      & sheep      & sofa       & train      & tv         & mAP       \\
\hline
scratch  &0  &  66.1  &  58.1  &  32.7  &  23.0  &  21.8  &  54.5  &  56.4  &  50.8  &  21.6  &  42.2  &  31.8 &  49.2   &  49.8  &  61.6  &  52.1  &  25.1  &  52.6  &  31.3  &  50.0  &  49.1  &  44.0  \\
{scratch (3 ensemble)}  &0  &  68.7  &  61.2  &  36.1  &  25.7  &  24.3  &  58.9  &  58.8  &  55.3  &  24.4  &  43.5  &  36.7 &  53.0   &  53.8  &  65.6  &  54.3  &  27.3  &  53.5  &  38.3  &  54.6  &  51.8  &  47.3  \\
unsup + ft & 1.5M  &  68.8  &  62.1  &  34.7  &  25.3  &  26.6  &  57.7  &  59.6  &  56.3  &  22.0  &  42.6  &  33.8 &  52.3   &  50.3  &  65.6  &  53.9  &  25.8  &  51.5  &  32.3  &  51.7  &  51.8  &  46.2  \\
unsup + ft & 5M  & 69.0 & 64.0 & 37.1 & 23.6 & 24.6 & 58.7 & 58.9 & 59.6 & 22.3 & 46.0 & 35.1 & 53.3 & 53.7 & 66.9 & 54.1 & 25.4 & 52.9 & 31.2 & 51.9 & 51.8 & 47.0  \\
unsup + ft & 8M  & 67.6 & 63.4 & 37.3 & 27.6 & 24.0 & 58.7 & 59.9 & 59.5 & 23.7 & 46.3 & 37.6 & 54.8 & 54.7 & 66.4 & 54.8 & 25.8 & 52.5 & 31.2 & 52.6 & 52.6 & 47.5  \\
unsup + ft (2 ensemble) &6.5M & 72.4 & 66.2 & 41.3 & 26.4 & 26.8 & 61.0 & 61.9 & 63.1 & 25.3 & 51.0 & 38.7 & 58.1 & 58.3 & 70.0 & 56.2 & \textbf{28.6} & 56.1 & 38.5 & 55.9 & 54.3 & 50.5 \\
unsup + ft (3 ensemble) & 8M & 73.4 & 67.3 & 44.1 & 30.4 & 27.8 & \textbf{63.3} & \textbf{62.6} & 64.2 & {27.7} & 51.1 & 40.6 & 60.8 & 59.2 & {71.2} & \textbf{58.5} & {28.2} & 55.6 & 39.4 & \textbf{58.0} & 56.1 & 52.0 \\
\hline
unsup + iterative ft & 5M  & 67.7 & 64.0 & 41.3 & 25.3 & 27.3 & 58.8 & 60.3 & 60.2 & 24.3 & 46.7 & 34.4 & 53.6 & 53.8 & 68.2 & 55.7 & 26.4 & 51.1 & 34.3 & 53.4 & 52.3 & 48.0 \\
\hline
RCNN 70K & &  72.7  &  62.9  &  49.3  &  31.1  &  25.9  &  56.2  &  53.0  &  70.0  &  23.3  &  49.0  &  38.0  &  69.5  &  60.1  &  68.2  &  46.4  &  17.5  &  57.2  &  46.2  &  50.8  &  54.1  &  50.1 \\
RCNN 70K (2 ensemble) & &  \textbf{75.3}  &  {68.3}  &  {53.1}  &  {35.2}  &  27.7  &  59.6  &  54.7  &  {73.4}  &  26.5  &  53.0  &  {42.2}  &  {73.1}  &  \textbf{66.1}  &  71.0  &  48.5  &  21.7  &  {59.2}  &  {50.8}  &  55.2  &  \textbf{58.0}  & {53.6} \\
{RCNN 70K (3 ensemble)}  &  &  74.6  &  \textbf{68.7}  &  \textbf{54.9}  &  \textbf{35.7}  &  29.4  &  61.0  &  54.4  &  \textbf{74.0}  &  \textbf{28.4}  &  \textbf{53.6}  &  \textbf{43.0} &  \textbf{74.0}   &  \textbf{66.1}  & \textbf{72.8}  &  50.3  &  20.5  &  \textbf{60.0}  &  \textbf{51.2}  &  57.9  & \textbf{ 58.0}  &  \textbf{54.4}  \\
RCNN 200K (big stepsize) & & 73.3 & 67.1 & 46.3 & 31.7 & \textbf{30.6} & 59.4 & 61.0 & 67.9 & {27.3} & {53.1} & 39.1 & 64.1 & 60.5 & 70.9 & 57.2 & 26.1 & 59.0 & 40.1 & 56.2 & 54.9 & 52.3\\
\end{tabular}
}
\label{tab:VOCAP}
\vspace{-0.2in}
\end{table*}

At the fine-tuning stage,  we change the output to $21$ classes and initialize the convolutional layers with our unsupervised pre-trained network. To fine-tune the network, we start with learning rate as $\epsilon =0.01$ and reduce the learning rate by a factor of 10 at every 80K iterations. The network is fine-tuned for 200K iterations. Note that for all the experiments, no bounding box regression is performed.

We compare our method with the model trained from scratch as well as using ImagNet pre-trained network. Notice that the results for VOC 2012 reported in RCNN~\cite{girshick14rcnn} are obtained by only fine-tuning on the train set without using the val set. For fair comparison, we fine-tuned the  ImageNet pre-trained network with VOC 2012 trainval set. Moreover, as the step size of reducing learning rate in RCNN~\cite{girshick14rcnn} is set to 20K and iterations for fine-tuning is 70K, we also try to enlarge the step size to 50K and fine-tune the network for 200K iterations. We report the results for both of these settings.

\textbf{Single Model.} We show the results in Table~\ref{tab:VOCAP}. As a baseline, we train the network from scratch on VOC 2012 dataset and obtain  $44\%$ mAP. Using our unsupervised network pre-trained with  1.5M pair of patches and then fine-tuned on VOC 2012, we obtain mAP of  $46.2\%$ (unsup+ft, external data = 1.5M). However, using more data, 5M  and 8M patches in pre-training and then fine-tune, we can achieve $47\%$ and $47.5\%$ mAP. These results indicate that our unsupervised network provides a significant boost as compared to the scratch network. More importantly, when more unlabeled data is applied, we can get better performance ($~3.5\%$ boost compared to training from scratch).

\textbf{Model Ensemble.} We also try combining different models using different sets of unlabeled data in pre-training. By ensembling two fine-tuned networks which are pre-trained using 1.5M and 5M patches, we obtained a boost of $3.5\%$ comparing to the single model, which is $50.5\%$(unsup+ft (2 ensemble)). Finally, we ensemble all three different networks pre-trained with different sets of data, whose size are 1.5M, 5M and 8M respectively. We get another boost and reach $52\%$ mAP (unsup+ft (3 ensemble)).

\textbf{Baselines.} We compare our approach with RCNN~\cite{girshick14rcnn} which uses ImageNet pre-trained models. Following the procedure in~\cite{girshick14rcnn}, we obtain $50.1\%$ mAP (RCNN 70K) by setting the step size to 20K and fine-tuning for 70K iterations. To generate a model ensemble, the CNNs are first trained on the ImageNet dataset separately, and then they are fine-tuned with the VOC 2012 dataset. The result of ensembling two of these networks is $53.6\%$ mAP (RCNN 70K (2 ensemble)). If we ensemble three networks, we get a mAP of $54.4\%$. For fair of comparison, we also fine-tuned the ImageNet pre-trained model with larger step size (50K) and more iterations (200K). The result is $52.3\%$ mAP (RCNN 200K (big stepsize)). Note that while ImageNet network shows diminishing returns with ensembling since the training data remains similar, in our case since every network in the ensemble looks at different sets of data, we get huge performance boosts.

\textbf{Exploring a better way to transfer learned representation.} Given our fine-tuned model using 5M patches in pre-training (unsup+ft, external = 5M), we use it to re-learn and re-adapt to the unsupervised triplet task. After that, the network is re-applied to fine-tune on VOC 2012. The final result for this single model is $48\%$ mAP (unsup + iterative ft), which is $1\%$ better than the initial fine-tuned network.

\textbf{Unsupervised network without fine-tuning:} We also perform object detection without fine-tuning on VOC 2012. We extract pool5 features using our  unsupervised-CNN and train SVM on top of it. We obtain mAP of $26.1\%$ using our unsupervised network (training with 8M data). The ensemble of two unsupervised-network (training with 5M and 8M data) gets mAP of $28.2\%$. As a comparison, Imagenet pre-trained network without fine-tuning gets mAP of $40.4\%$.

\vspace{-0.175in}
\subsubsection{Surface Normal Estimation}
\vspace{-0.085in}
\begin{table}
\centering
\caption{{\footnotesize Results on NYU v2 for per-pixel surface normal estimation, evaluated over valid pixels.}}
\label{tab:3dresults}
\small
\begin{tabular}{@{}l@{ }c@{ }c@{ }c@{ }c@{ }c@{ }c} \toprule
        & \multicolumn{2}{c}{(Lower Better)} & \multicolumn{3}{c}{(Higher Better)} \\
        & ~Mean~~~ & Median & $11.25^{\circ}$ & $22.5^{\circ}$ & $30^{\circ}$ \\ \midrule

scratch                 & 38.6      &  26.5  & 33.1  & 46.8      & 52.5 \\
 unsup + ft                & 34.2      &  21.9  & 35.7  & 50.6      & 57.0 \\
ImageNet + ft             & 33.3      &  20.8  & 36.7  & 51.7      & 58.1 \\
UNFOLD \cite{Fouhey14c}     & 35.1      & 19.2      & 37.6      & 53.3  & 58.9 \\
Discr. \cite{Ladicky14b}    & 32.5      & 22.4      & 27.4      & 50.2  & 60.2 \\
3DP (MW) \cite{Fouhey13a}   & 36.0      & 20.5      & 35.9      & 52.0  & 57.8 \\ \bottomrule
\end{tabular}
\vspace{-0.15in}
\end{table}

To illustrate that our unsupervised representation can be generalized to different tasks, we adapt the unsupervised CNN to the task of surface normal estimation from a RGB image. In this task, we want to estimate the orientation of the pixels. We perform our experiments on the NYUv2 dataset~\cite{Silberman12}, which includes 795 images for training and 654 images for testing. Each image is has corresponding depth information which can be used to generate groundtruth surface normals. For evaluation and generating the groundtruth, we adopt the protocols introduced in~\cite{Fouhey13a} which is used by different methods~\cite{Fouhey13a,Ladicky14b,Fouhey14c} on this task.

To apply deep learning to this task, we followed the same form of outputs and loss function as the coarse network mentioned in~\cite{deep3d15}. Specifically, we first learn a codebook by performing k-means on surface normals and generate 20 codewords. Each codeword represents one class and thus we transform the problem to 20-class classification for each pixel. Given a $227 \times 227$ image as input, our network generates surface normals for the whole scene. The output of our network is $20 \times 20$ pixels, each of which is represented by a distribution over 20 codewords. Thus the dimension of output is $20 \times 20 \times 20 =8000$.

The network architecture for this task is also based on the AlexNet. To relieve over-fitting, we only stack two fully connected layers with $4096$ and $8000$ neurons on the pool5 layer. During training, we initialize the network with the unsupervised pre-trained network (single network using 8M external data). We use the same learning rate $1.0 \times 10^{-6}$ as ~\cite{deep3d15} and fine-tune with 10K iterations given the small number of training data. Note that unlike~\cite{deep3d15}, we do not utilize any data from the videos in NYU dataset for training.

\begin{figure}
\includegraphics[width=0.47\textwidth]{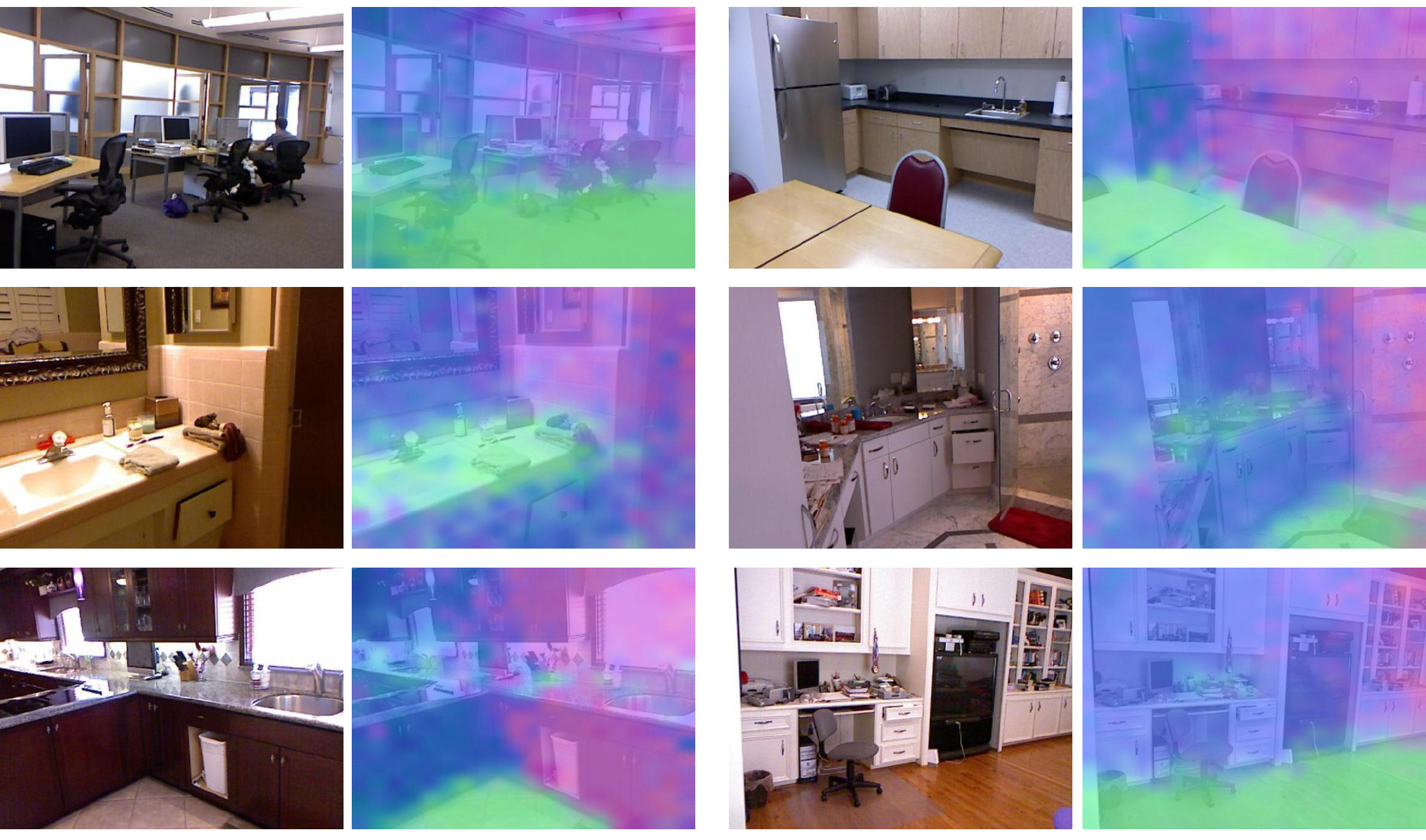}
\centering
\caption{\label{fig:3d}
Surface normal estimation results on NYU dataset. For visualization, we use green for horizontal surface, blue for facing right and red for facing left, i.e., \textcolor{blue}{blue $\rightarrow$ X}; \textcolor{green}{green $\rightarrow$ Y}; \textcolor{red}{red $\rightarrow$ Z.}}
\vspace{-0.16in}
\end{figure}

For comparison, we also trained networks from scratch as well as using ImageNet pre-trained. For evaluation, we report mean and median error (in degrees). We also report percentage of pixels with less than 11.25, 22.5 and 30 degree errors. We show our qualitative results in in Figure~\ref{fig:3d}. and quantitative results in Table~\ref{tab:3dresults}. Our approach (unsup + ft) is significantly better than network trained from scratch and comes very close to Imagenet-pretrained CNN ($\sim 1\%$).

\vspace{-0.10in}
\section{Discussion and Conclusion}
\vspace{-0.10in}
We have presented an approach to train CNNs in an unsupervised manner using videos. Specifically, we track millions of patches and learn an embedding using CNN that keeps patches from same track closer in the embedding space as compared to any random third patch. Our unsupervised pre-trained CNN fine-tuned using VOC training data outperforms CNN trained from scratch by $3.5\%$. An ensemble version of our approach outperforms scratch by $4.7\%$ and comes tantalizingly close to an Imagenet-pretrained CNN (within $2.5\%$). We believe this is an extremely surprising result since until recently semantic supervision was considered a strong requirement for training CNNs. We believe our successful implementation opens up a new space for designing unsupervised learning algorithms for CNN training.

{\footnotesize
\noindent {\bf Acknowledgement}: This work was partially supported by ONR MURI N000141010934 and NSF IIS 1320083. This material was also based on research partially sponsored by DARPA under agreement number FA8750-14-2-0244. The U.S. Government is authorized to reproduce and distribute reprints for Governmental purposes notwithstanding any copyright notation thereon. The views and conclusions contained herein are those of the authors and should not be interpreted as necessarily representing the official policies or endorsements, either expressed or implied, of DARPA or the U.S. Government. The authors would like to thank Yahoo! and Nvidia for the compute cluster and GPU donations respectively.
}

\vspace{-0.1in}

{\footnotesize
\bibliographystyle{ieee}
\bibliography{local}
}

\end{document}